\documentclass[letterpaper, 10 pt, conference]{ieeeconf}  

\IEEEoverridecommandlockouts                              
\usepackage{amsmath}    			
\usepackage{amssymb}
\usepackage{amsfonts}
\usepackage{nccmath}

\usepackage{graphicx}  				

\usepackage[croatian]{babel}  
\usepackage[utf8]{inputenc}  	
\usepackage[T1]{fontenc}
\usepackage{ae,aecompl}     	

\usepackage{microtype}				

\usepackage{subfig}

\usepackage{tabularx}
\usepackage{booktabs}
\usepackage{enumerate}				
\usepackage{algorithm2e}			
\usepackage{todonotes}				
\usepackage{dirtree}					
\usepackage{hyperref}					
\usepackage{color}
\usepackage{cite}
\graphicspath{{./figures/}}
\usepackage{graphicx}

\usepackage{amsthm}
\usepackage{bm}

\title{\LARGE \bf
 Error State Extended Kalman Filter Multi-Sensor Fusion for Unmanned Aerial Vehicle Localization in GPS and Magnetometer Denied Indoor Environments
}
\author{Lovro Marković, Marin Kovač, Robert Milijas, Marko Car, Stjepan Bogdan
	\thanks{Authors are with Faculty of Electrical and Computer Engineering,
        University of Zagreb, 10000 Zagreb, Croatia
        {\tt\small (lovro.markovic, marin.kovac, robert.milijas, marko.car, matko.orsag, stjepan.bogdan) at fer.hr}}}%

\makeatletter
\newcommand{\removelatexerror}{\let\@latex@error\@gobble}

\makeatother

\begin{document}
\maketitle

\thispagestyle{empty}
\pagestyle{empty}

\begin{abstract}
This paper addresses the issues of unmanned aerial vehicle (UAV) indoor navigation, specifically in areas where GPS and magnetometer sensor measurements are unavailable or unreliable. The proposed solution is to use an error state extended Kalman filter (ES -EKF) in the context of multi-sensor fusion.  Its implementation is adapted to fuse measurements from multiple sensor sources and the state model is extended to account for sensor drift and possible calibration inaccuracies. Experimental validation is performed by fusing IMU data obtained from the PixHawk 2.1 flight controller with pose measurements from LiDAR Cartographer SLAM, visual odometry provided by the Intel T265 camera and position measurements from the Pozyx UWB indoor positioning system. The estimated odometry from ES-EKF is validated against ground truth data from the Optitrack motion capture system and its use in a position control loop to stabilize the UAV is demonstrated.
\end{abstract}
\section{Introduction}
\label{sec:introduction}

Unmanned aerial vehicles (UAVs) are becoming increasingly popular as their capabilities such as autonomy, computing power, and payload weight increase. Thanks to their increased payload weight, UAVs can adopt a diverse sensor suite including Light Detection and Ranging (LiDAR) sensors, depth and multispectral cameras, GPS and ultra wideband (UWB) receivers, magnetometer modules, etc. All of these sensors provide important information for localization and stabilization of the UAV. Depending on where the UAV is deployed, reliable and consistent localization is an ongoing problem \cite{Alatise2020}. In the context of fusing multiple measurements, different sensor combinations may prove more useful than others. In this paper, the focus is on indoor environments where GPS receivers cannot provide an accurate position lock and magnetometers report inconsistent measurements due to possible electromagnetic interference in the vicinity. 

State estimation techniques related to sensor fusion are often formulated as filtering and optimization problems. The optimization methods \cite{Ding2021},\cite{Strasdat2012}, are more precise, but often require a significant amount of computational resources. Due to the scope of the state estimation field, this paper focuses on filtering techniques, mostly based on the extended Kalman filter (EKF) or its variants, in the context of multi-sensor fusion. In \cite{Yang2005}, a particle filter (PF) is used in a land navigation system for sensor fusion and a Monte-Carlo variant of PF is presented in \cite{PerezGrau2017},\cite{Carrasco2021} for UAV localization in GPS denied environments. A sigma-point Kalman filter (SPKF) is used for integrated navigation purposes and GPS/IMU fusion in \cite{vanderMerwe2004} and \cite{WENDEL2006} respectively. The authors in  \cite{Kelly2010} utilize an unscented Kalman filter (UKF) in a self-calibrating visual-inertial sensor fusion framework and \cite{You2020} presents a UKF-based fusion of UWB and inertial measurements for indoor UAV localization.
A novel multi-state constraint Kalman filter (MSCKF) is presented in \cite{Lee2020} used for the fusion of GPS and visual-inertial odometry (VIO) measurements. Although suboptimal, regular EKF implementations are still widely used techniques for state estimation in the context of multi-sensor fusion \cite{Ntzi2010},\cite{Weiss2012},\cite{Lpez2015}. The work presented in \cite{Lpez2015} is outlined as visual and SLAM sensors are incorporated into the state estimation for indoor UAV localization.  
\begin{figure}[t]
	\centering
	\includegraphics[width=\columnwidth]{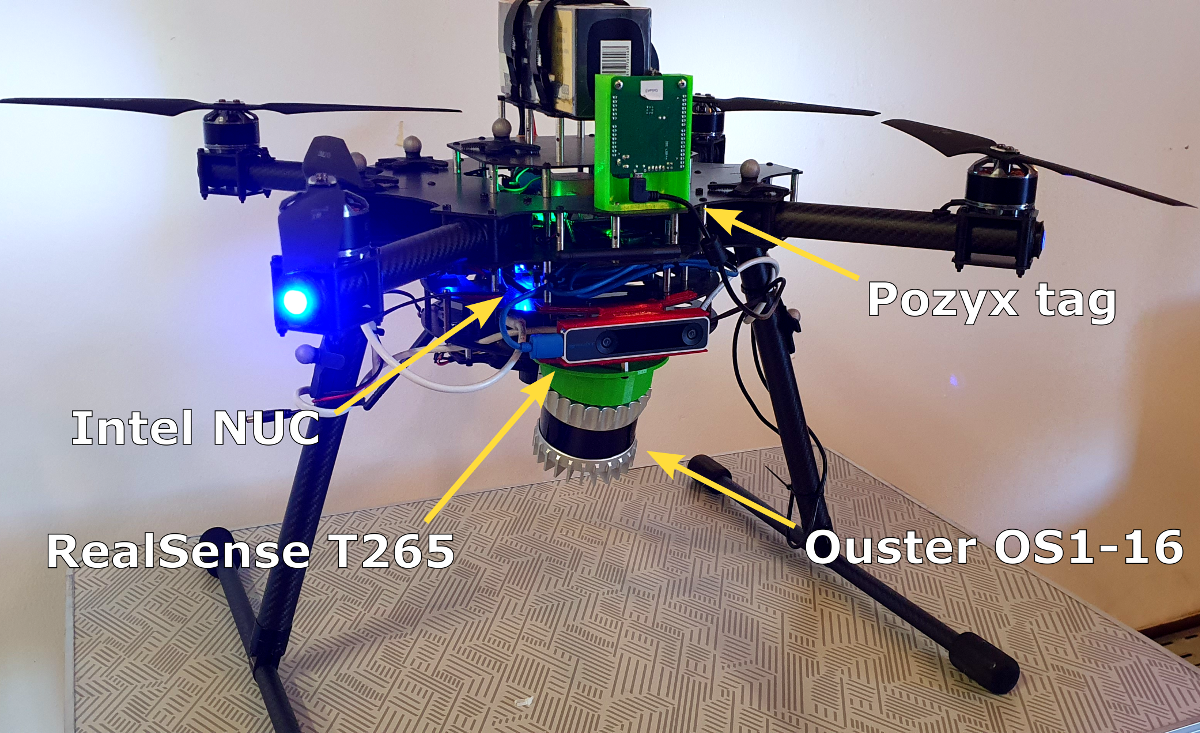}
	\caption{The figure shows a custom built UAV used for indoor localization experiments. Reflective markers are attached to the top of the UAV to provide ground truth measurements from the Optitrack motion capture system. An Ouster OS1-16 high-resolution LiDAR sensor, when used with Cartographer SLAM, provides pose sensing capabilities. A Pozyx UWB accurate positioning sensor is attached to the top of the UAV and provides position measurements. Finally, an Intel RealSense T265 camera provides visual odometry information.}
	\label{fig:hawk_uav}
	\vspace{-0.7cm}
\end{figure}
\\The error state extended Kalman filter (ES-EKF) relies on linear error state dynamics for optimal prediction and update of the error state covariance. A comparative analysis between the classic EKF and the ES-EKF performed in \cite{Liu2019},\cite{Madyastha2011},\cite{Guo2019},\cite{Wonkeun2019} shows that the ES-EKF is robust to a variety of aircraft maneuvers as well as imperfect tuning of the sensor noise covariance. Moreover, it ensures numerical stability and allows the quaternions to be treated in their minimal representation. The error state is generally small in magnitude which prevents singularities and gimbal lock from occuring. Finally, the ES-EKF is more suitable for UAV localization and stabilization due to its higher computational efficiency.
An application of the ES-EKF filter in multi-sensor state estimation for underwater vehicle localization is presented in \cite{Shaukat2021}. Other notable work includes \cite{Lynen2013}, in which the authors apply an iterative extended Kalman filter (I-EKF) based on an error state representation in a multi-sensor fusion framework and demonstrate the approach for the case of outdoor UAV localization. It also allows automatic estimation of the sensor transformations and fusing absolute as well as relative measurements. The work in \cite{Bulunseechart2018} also utilizes I-EKF for localization in GNSS-limited environments.\\
In this paper, a solution for UAV localization in GPS and magnetometer denied environments is proposed. Due to its robustness and computational stability, an ES-EKF is used for reliable state estimation in the context of multi-sensor fusion. The work presented in \cite{sola2017} gives an overview of quaternion kinematics and provides an intuitive insight into the technique of error state filtering and serves as a starting point for the approach proposed in this paper. The filter implementation is further extended to allow for online drift estimation. Visual and LiDAR SLAM, commonly used as pose sensing methods in the context of indoor localization, are known to accumulate drift error over time, hence the motivation to consider this in state estimation. \\
The main contributions of this work are as follows. An ES-EKF is adapted to allow fusion of multiple measurement sources (position, orientation and velocity) for indoor UAV localization not supported by GPS or magnetometer measurements. The filter state is also extended to allow position and orientation drift estimation for each sensor. In addition, an arbiter for sensor measurements is introduced to eliminate outliers before the fusion process.  Finally, the approach is validated in an indoor laboratory using motion capture cameras, with the on-board sensor suite containing both visual and LiDAR SLAM measurements as well as an UWB positioning anchor. The estimated UAV state is also used for stabilization in a position control loop. \\
The paper is organized as follows. In  Sec. \ref{sec:system_kinematics}, an overview of nominal and error state kinematics is given, along with the equations for the filter prediction. The approach to fusing measurements from multiple sources using the ES-EKF formulation is presented in Sec. \ref{sec:sensor_fusion}. In Sec. \ref{sec:experiments}, an experimental validation of the described approach is performed in an indoor laboratory setup with measurements obtained from a variety of sensors suitable for indoor navigation. Finally, the paper is concluded in Sec. \ref{sec:conclusion}.
\section{System Kinematics}
\label{sec:system_kinematics}

In this section, the notions of the nominal and the error state are presented as core concepts of the error state filter formulation. The equations used for the prediction step in the ES-EKF are shown. Finally, the true state is given as a combination of the nominal and the error state.
\subsection{Nominal State}
The nominal state kinematics are obtained by integrating high-frequency IMU data $\textbf{u}_{mv} = \left[ \textbf{a}_{mv}, \bm{\omega}_{mv} \right]^T$ containing measured acceleration and angular velocity values respectively. The nominal state kinematics are presented as follows:
\begin{align}
    \textbf{p} &= \textbf{p} + \textbf{v} \Delta t + \frac{1}{2} \left( \text{R} (\textbf{a}_{mv} - \textbf{a}_b) + \textbf{g} \right) \Delta t^2, \\
    \textbf{v} &= \textbf{v} + \left( R  (\textbf{a}_{mv} - \textbf{a}_b) + \textbf{g} \right) \Delta t, \\
    \textbf{q} &= \textbf{q} \cdot \textbf{q}\{(\bm{\omega}_{mv} - \bm{\omega}_b)\Delta t\}, \\
    \textbf{a}_b &= \textbf{a}_b, \\
    \bm{\omega}_b &= \bm{\omega}_b, \\
    \textbf{g} &= \textbf{g}, \\
    \textbf{p}_i &= \textbf{p}_i, \quad \forall i \in \text{S}_{drift} \\
    \textbf{q}_i &= \textbf{q}_i, \quad \forall  i \in \text{S}_{drift},
\end{align}
where \textbf{q}\{\textbf{x}\} is a quaternion representation of the angle-axis vector \textbf{x}, $\text{S}_{drift}$ is the set of all available sensors, $\textbf{a}_b$ and $\bm{\omega}_b$ are acceleration and angular velocity biases respectively. Estimated drift transformation for each individual sensor is denoted by position $\textbf{p}_i$ and orientation $\textbf{q}_i$.
\subsection{Error State}
The errors accumulated in the nominal state due to the omitted noise terms and model inaccuracies are collected in the error state. It has the following notation:
\begin{equation}
    \delta \textbf{x} = [\delta \textbf{p}, \delta \textbf{v}, \delta \bm{\theta}, \delta \textbf{a}_b, \delta \bm{\omega}_b, \delta \textbf{g}, \delta \textbf{p}_i, \delta \bm{\theta}_i]^T.
\end{equation}
The error state angle-axis notation of the UAV orientation  and the sensor drift orientation, $\delta \bm{\theta}$ and $\delta \bm{\theta}_i$ respectively, is used as the minimum rotation representation. This does not expose the system to the risk of singularities or gimbal-lock, due to the small magnitude of the error state. The same considerations apply to explaining the accuracy of the linear nature of the system kinematics of the error state. Small magnitudes of the error state allow the second-order terms to be omitted without a significant loss of precision. The model of the error state kinematics with included system noise $\textbf{w}$ is presented as follows:
\begin{gather}
    \delta x = \text{F}_x \delta x + \text{F}_w \textbf{w}. \\
    \text{P} = \text{F}_x \text{P} {\text{F}_x}^T + \text{F}_w \text{Q}_w {\text{F}_w}^T.
\end{gather}

The complete derivation of $\text{F}_x$ is found in \cite{sola2017}. Since the error state, in  this paper, is extended to include sensor drift estimation $\text{F}_x$ is also extended as follows:
\begin{gather}
    \text{F}_x = \frac{\partial f}{\partial \delta x} = \nonumber\\
    \medmath{\begin{bmatrix}
        \text{I}_3 & \text{I}_3 \Delta t & 0_3 & 0_3 & 0_3 & 0_3 & 0_3 & 0_3 \\
        0_3 & \text{I}_3 & - \text{R}\cdot\text{A} & - \text{R} \Delta t & 0_3 & \text{I}_3 \Delta t & 0_3 & 0_3 \\
        0_3 & 0_3 & \text{R}^T\{\bm{\omega}\} & 0_3 & - \text{I}_3 \Delta t & 0_3 & 0_3 & 0_3 \\
        0_3 & 0_3 & 0_3 & \text{I}_3 & 0_3 & 0_3 & 0_3 & 0_3 \\
        0_3 & 0_3 & 0_3 & 0_3 & \text{I}_3 & 0_3 & 0_3 & 0_3 \\
        0_3 & 0_3 & 0_3 & 0_3 & 0_3 & \text{I}_3 & 0_3 & 0_3 \\
        0_3 & 0_3 & 0_3 & 0_3 & 0_3 & 0_3 & \text{I}_3 & 0_3 \\
        0_3 & 0_3 & 0_3 & 0_3 & 0_3 & 0_3 & 0_3 & \text{I}_3 \\
    \end{bmatrix}} \\
    \quad \text{A}=\left[(\textbf{a}_{mv} - \textbf{a}_b) \right]_\times \Delta t, \quad \bm{\omega} = (\bm{\omega}_{mv} - \bm{\omega}_b)\Delta t, \\ 
    \text{F}_w = \frac{\partial f}{\partial \textbf{w}} =
    \begin{bmatrix}
        0_3 & 0_3 & 0_3 & 0_3 \\
        \text{I}_3 & 0_3 & 0_3 & 0_3 \\
        0_3 & \text{I}_3 & 0_3 & 0_3 \\
        0_3 & 0_3 & \text{I}_3 & 0_3 \\
        0_3 & 0_3 & 0_3 & \text{I}_3 \\
        0_3 & 0_3 & 0_3 & 0_3 \\
        0_3 & 0_3 & 0_3 & 0_3 \\
        0_3 & 0_3 & 0_3 & 0_3
    \end{bmatrix}, \\
    \medmath{\text{Q}_{w} = 
    \begin{bmatrix}
        \bm{\sigma}_v^2\Delta t ^ 2 \text{I}_3 & 0_3 & 0_3 & 0_3 \\
        0_3 & \bm{\sigma}_{\theta}^2 \Delta t ^ 2 \text{I}_3 & 0_3 & 0_3\\
        0_3 & 0_3 & \bm{\sigma}_a^2 \Delta t ^ 2 \text{I}_3 & 0_3 \\
        0_3 & 0_3 & 0_3 & \bm{\sigma}_{\omega}^2 \Delta t ^ 2 \text{I}_3
    \end{bmatrix}},
\end{gather}
where R\{\textbf{x}\} is a matrix rotation representation of the \textbf{x} angle-axis rotation, $\bm{\sigma}^2_v$, $\bm{\sigma}^2_{\theta}$, $\bm{\sigma}^2_a$ and $\bm{\sigma}^2_{\omega}$ are velocity, orientation, acceleration bias and angular velocity bias variances respectively. Skew-symmetric operator is defined as:
\begin{equation}
    [\textbf{x}]_\times : \mathbf{R}^3 \longrightarrow \mathfrak{so}(3).
\end{equation}

\section{Multi-Sensor Fusion}
\label{sec:sensor_fusion}

This section shows how the correction step for the ES-EKF implementation is designed in the context of multi-sensor fusion. Furthermore, the additions addressing the sensor drift position and orientation are outlined. Baseline Kalman filter correction step is presented as follows:
\begin{align}
    \text{K} &= \text{P} \text{H}^T ( \text{H} \text{P} \text{H}^T + \text{V})^{-1}, \quad \text{V} = \bm{\sigma}_s^2 \text{I} \\
    \widehat{\delta \textbf{x}} &= \text{K} ( \textbf{y}_{mv} - h(\widehat{\textbf{x}_t})), \label{eqn:inovation}\\
    \text{P} &= (\text{I}_3 - \text{K} \text{H}) \text{P}, \\
    h(\widehat{\textbf{x}_t}) &= 
    \begin{cases}
        \text{R}\{\textbf{q}_i\} \widehat{\textbf{p}_t} + \textbf{p}_i, \quad \text{position sensor}, \\
        \textbf{q}_i \cdot \widehat{\textbf{q}_t}, \quad \text{orientation sensor}, \\
        \widehat{\textbf{x}_t}, \quad \text{otherwise}.
    \end{cases}
\end{align}
True state $\widehat{\textbf{x}_t}$ is the current best estimate of the UAV state. It is expressed as a combination of the nominal state and error state as follows:
\begin{align}
     &\widehat{\textbf{p}_t} = \textbf{p} + \widehat{\delta \textbf{p}}, \\
    &\widehat{\textbf{v}_t} = \textbf{v} + \widehat{\delta \textbf{v}}, \\
    &\widehat{\textbf{q}_t} = \textbf{q} \cdot \delta \textbf{q}\{\widehat{\delta \bm{\theta}}\}, \\
    &\widehat{\textbf{a}_{b,t}} = \textbf{a}_b + \widehat{\delta \textbf{a}_b}, \\
    &\widehat{\bm{\omega}_{b,t}} = \bm{\omega}_b, + \widehat{\delta \bm{\omega}_b}\\
    &\widehat{\textbf{g}_t} = \textbf{g} + \widehat{\delta \textbf{g}}, \\
    &\widehat{\textbf{p}_{i,t}} = \textbf{p}_i + \widehat{\delta \textbf{p}_i}, \quad \forall i \in \text{S}_{drift} \\
    &\widehat{\textbf{q}_{i,t}} = \textbf{q}_i \cdot \delta \textbf{q} \{\widehat{\delta \bm{\theta}_i }\}, \quad \forall  i \in \text{S}_{drift},
\end{align}
The linearized matrix H is defined as follows:
\begin{equation}
    \text{H} = \frac{\partial h}{\partial \delta \textbf{x}} =
    \frac{\partial h}{\partial \textbf{x}_t} \frac{\partial \textbf{x}_t}{\partial \delta \textbf{x}}.
\end{equation}
It assumes different forms depending on the type of measurement received: position, orientation or velocity. For position measurements with drift estimation the Jacobian H components are derived as follows:
\begin{gather}
    \frac{\partial h}{\partial \textbf{x}_t} = 
    \begin{bmatrix}
    \text{R}\{\textbf{q}_i\} & 0_3 & 0_3 & 0_3 & 0_3 & 0_3 & \text{I}_3 & 0_3
    \end{bmatrix}, \label{eqn:jacobian_true}\\
    \frac{\partial \textbf{x}_t}{\partial \delta \textbf{x}} = \nonumber\\
    \medmath{\begin{bmatrix}
    \text{I}_3 & 0_3 & 0_3 & 0_{3\times12} & 0_{3}\\
    0_3 & \text{I}_3 & 0_3 & 0_{3\times12} & 0_3 \\
    0_{4\times3} & 0_{4\times3} & \frac{\partial (\textbf{q}\cdot\delta \textbf{q}\{\widehat{\delta \bm{\theta}}\})}{\partial \delta \bm{\theta}} & 0_{4\times12} & 0_{4\times3} \\
    0_{12\times3} & 0_{12\times3} & 0_{12\times3} & \text{I}_{12\times12} & 0_{12\times3}  \\
    0_{4\times3} & 0_{4\times3} & 0_{4\times3} & 0_{4\times3} & \frac{\partial (\textbf{q}_i\cdot \delta \textbf{q} \{\delta \bm{\theta}_i \})}{\partial \delta \bm{\theta}_i}
    \end{bmatrix}}, \label{eqn:jacobian_deltax}\\
    where \quad \frac{\partial (\textbf{q}\cdot\delta \textbf{q}\{\delta \bm{\theta}\})}{\partial \delta \bm{\theta}} = \frac{1}{2}
    \begin{bmatrix}
        - q_x & - q_y & - q_z \\
        q_w & - q_z & q_y \\
        q_z & q_w & - q_x \\
        -q_y & q_x & q_w
    \end{bmatrix}\label{eqn:partial_theta}.
\end{gather}
Complete proof of Eq. \ref{eqn:partial_theta} is found in \cite{sola2017}. When fusing other types of measurements the difference occurs in Eq. \ref{eqn:jacobian_true}. For orientation and velocity measurements the Jacobian component is defined as:
\begin{gather}
     \medmath{\frac{\partial h}{\partial \textbf{x}_t} = 
    \begin{bmatrix}
    0_3 & 0_3 & \text{I}_3 & 0_3 & 0_3 & 0_3 & 0_3 & \text{R}^T\{\textbf{q}_{mv}\}
    \end{bmatrix}}, \\
     \frac{\partial h}{\partial \textbf{x}_t} = 
    \begin{bmatrix}
    0_3 & \text{I}_3 & 0_3 & 0_3 & 0_3 & 0_3 & 0_3 & 0_3
    \end{bmatrix},
\end{gather}
respectively. Finally, the orientation measurement innovation term is calculated in its angle-axis representation of as follows:
\begin{equation}
    \textbf{e}_{\theta} = \left[ \text{log} \left( R^T \{ \textbf{q}_i \cdot \widehat{\textbf{q}_t} \} R \{ \textbf{q}_{mv} \} \right) \right] ^\vee, \label{eqn:orientation_error}
\end{equation}
where log and \textit{vee} maps are defined as:
\begin{align}
    \text{log} &: \text{SO}(3) \longrightarrow \mathfrak{so}(3), \\
    \left[ R \right] ^ \vee &: \mathfrak{so}(3) \longrightarrow \mathbb{R}^3,
\end{align}
respectively.

Sensor fault detection is also addressed prior to the observation fusion. The state estimation methods behind each sensor are inherently complex and error-prone. Therefore, to ensure that the filter fuses the healthiest measurements, an arbiter is implemented to eliminate potential outliers. A component-wise innovation limit test is introduced for each type of measurement. Since a single sensor can provide reliable position measurements in the horizontal plane while the vertical axis could be highly noisy, limits of different magnitudes may be required for each axis. The initial values for the arbiter limits are chosen as the maximum in-flight deviation from the mean and are further empirically tuned.
\section{Experimental Results}
\label{sec:experiments}

In this section, the indoor testing environment is presented along with the UAV platform and its components. Furthermore, the sensor data and the state estimation results are compared and analysed to verify the proposed approach.
\begin{figure}[h]
\subfloat[Position comparison along the X-axis.]{%
  \includegraphics[width=\columnwidth]{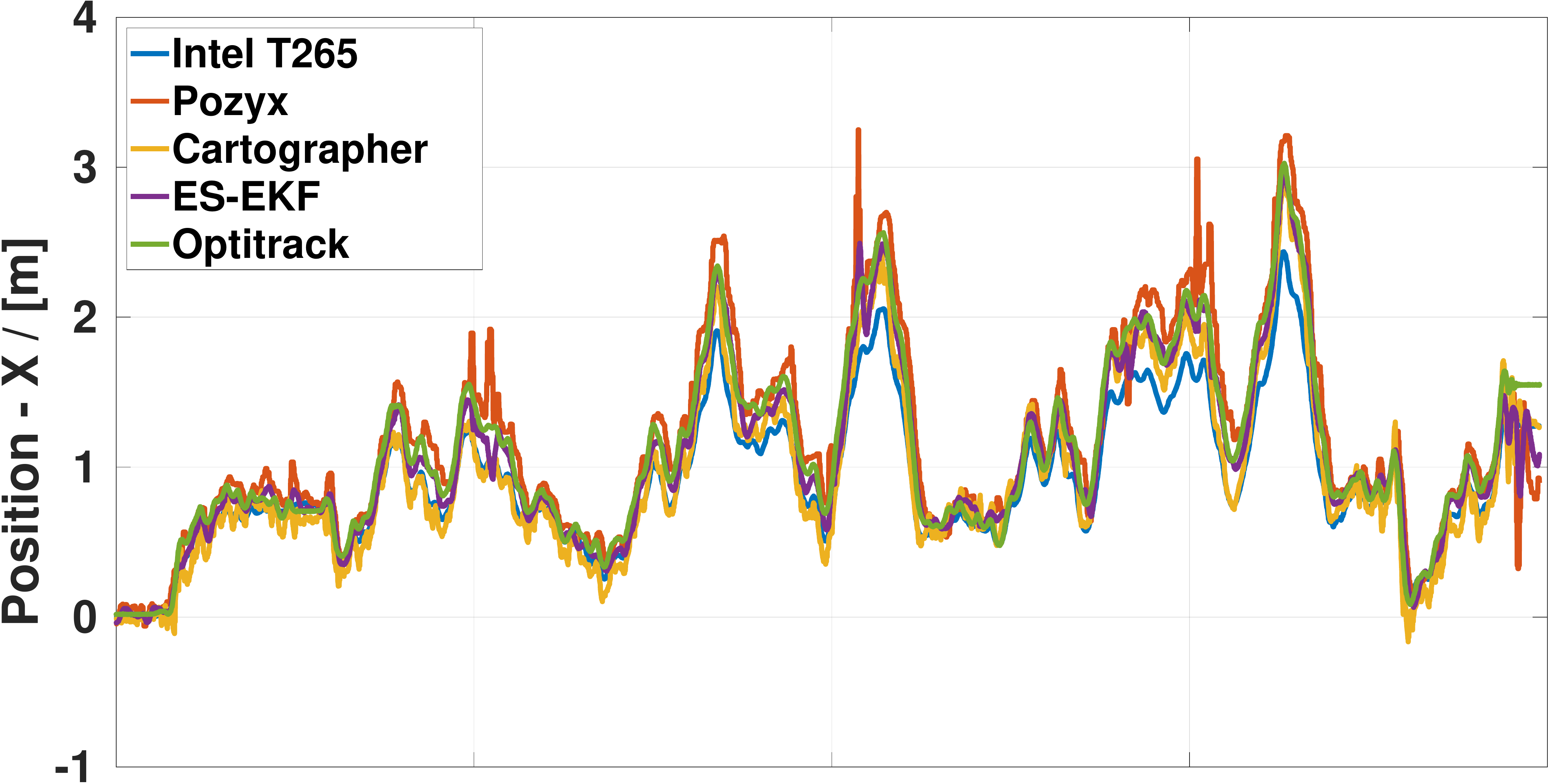}%
}\\
\subfloat[Position comparison along the Y-axis.]{%
  \includegraphics[width=\columnwidth]{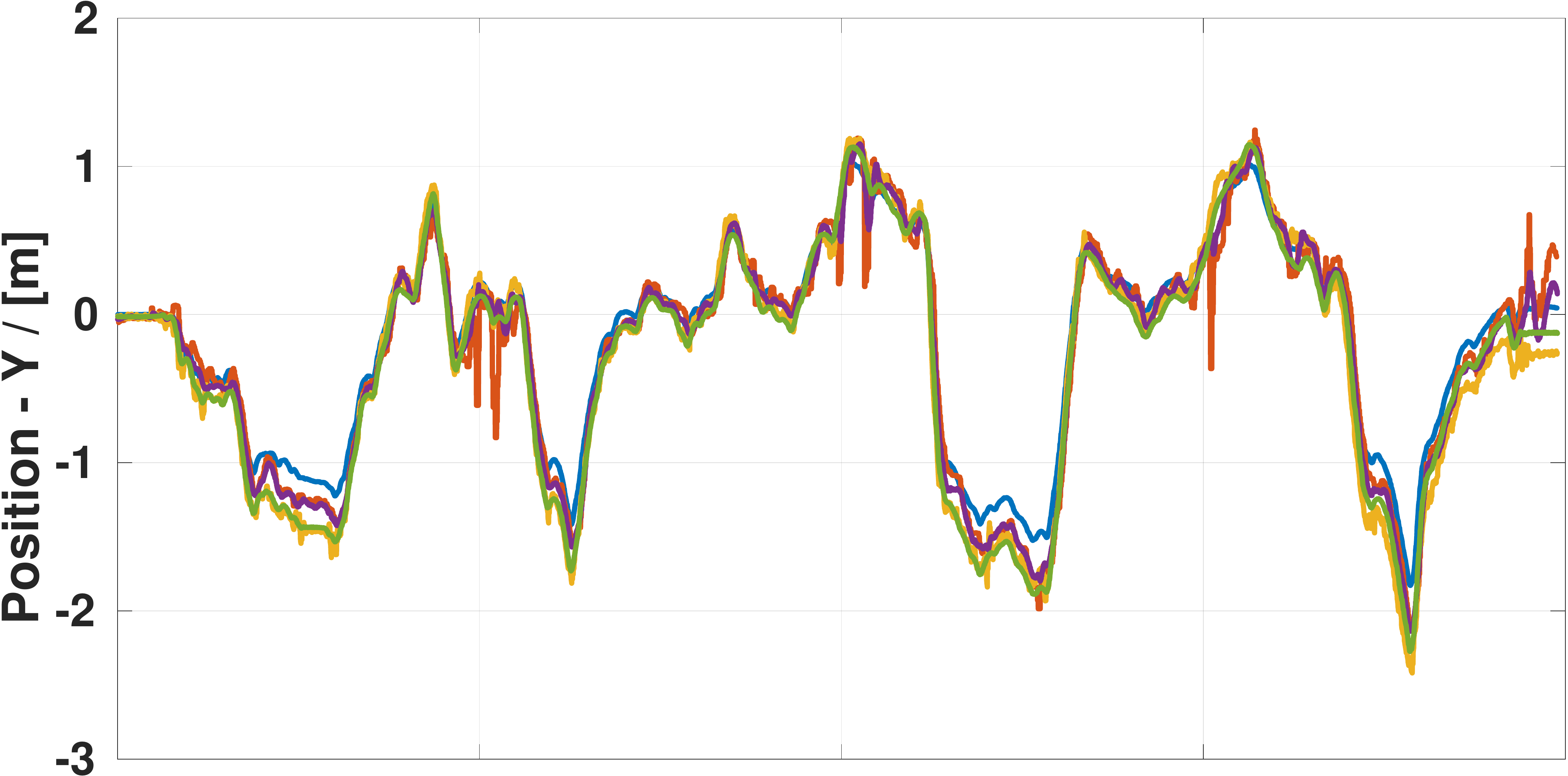}%
}\\
\subfloat[Position comparison along the Z-axis.]{%
  \includegraphics[width=\columnwidth]{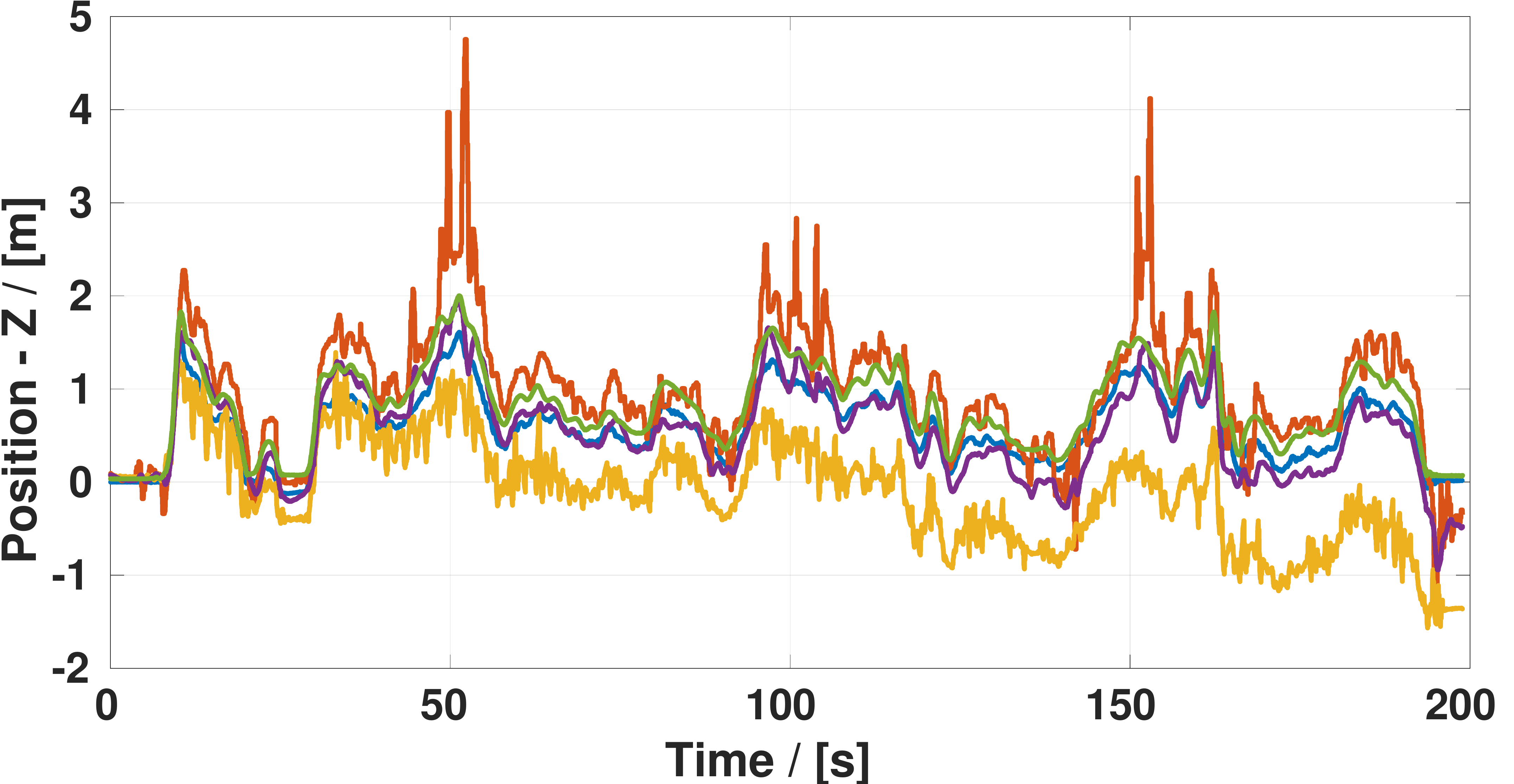}%
}
\caption{These figures show a comparison of position estimates along all three axes between Intel Realsense T265 VIO, Pozyx UWB positioning system, Cartographer SLAM, ES-EKF multi-sensor fusion and Optitrack motion capture system. The sensor coordinate frames are aligned wrt. each other with the estimated drift component removed, as explained in Sec. \ref{subsec:ekf_config}.}
\vspace{-0.5cm}
\label{fig:position_comparison}
\end{figure}
\subsection{Hardware Setup}
A custom-built UAV platform, shown in Fig. \ref{fig:hawk_uav} assembled by the \textit{Kopterworx} company, serves as a base for indoor localization experiments. The PixHawk 2.1 autopilot with ArduCopter firmware resides on a ProfiCNC/HEXKore  high-current power board. An Intel NUC on-board computer runs the entire UAV flight stack along with the ES-EKF multi-sensor fusion implementation within the ROS Melodic environment on Ubuntu Linux 18.04. The on-board computer is connected to the PixHawk autopilot via serial communication. This enables on-board control of the UAV using ROS, as well as receiving information about the UAV and the flight controller's sensors. IMU measurements obtained from PixHawk through the MAVRos interface are indispensable pieces of information for the proposed state estimation system. Taking in consideration that PixHawk provides three IMU sensors, the risk of failure is negligible. \\
There are three sensors mounted on the UAV that are used for sensor fusion. An Ouster OS1-16 high-resolution LiDAR sensor is used in combination with Cartographer SLAM \cite{Hess2016} to provide pose measurements. Previous research \cite{MilijasICUAS2021} has shown that Cartographer SLAM can be used as a pose sensor for UAV localization and stabilization in outdoor environments. In addition, LiDAR SLAM is also capable of producing high quality environment maps that are useful for obstacle navigation, interaction with the environment etc. especially when combined with a robust state estimation system to stabilize the UAV. The second sensor is the Pozyx UWB positioning system, the only sensor used that does not accumulate drift. The third sensor used is an Intel RealSense T265 camera that provides VIO information. The visual SLAM runs directly on the camera, freeing up some of the computing power for the on-board computer. \\
The indoor laboratory setup, as seen in the video attachment \cite{videoPlaylist}, uses the Optitrack motion capture system to provide ground truth measurements used for validation purposes only. Although the paper claims to provide robust localization in magnetometer denied environments, it is difficult to provide such an elaborate laboratory setup. Therefore, to mimic such conditions, all internal and external magnetometers on the UAV have been disabled or removed. 

\subsection{ES-EKF Configuration}
\label{subsec:ekf_config}
The parameter set for the ES-EKF is chosen as follows. Acceleration and angular velocity bias variances are obtained from the IMU specifications data sheet. The velocity and orientation process variances are tuned to balance the delay and noise of the filter estimate. The initial values for the sensor variance parameters are determined by observing signal-to-noise ratio of the raw sensor measurements. Each variance vector is further tuned based on \textit{a priori}  sensor measurement knowledge, e.g. Pozyx and Cartographer position estimates are known to be noisy along the Z-axis. \\
Furthermore, each sensor's initial measurement is considered its origin unless the the sensor does not measure the orientation, in which case a predefined rotation is used. Since the filter is able to compensate for transformation inaccuracies online, small errors in the initial sensor transformations are irrelevant. For accuracy reasons, all position and orientation measurements have their respective estimated drift removed from the measurement as follows:
\begin{gather}
    \textbf{p}_{no \, drift} = \text{R}^{-1}\{\textbf{q}_i\} \left( \textbf{p}_{mv} - \textbf{p}_i \right) \\
    \textbf{q}_{no \, drift} = \textbf{q}_i ^ {-1} \cdot \textbf{q}_{mv}
\end{gather}

\begin{figure}[t]
	\centering
	\includegraphics[width=\columnwidth]{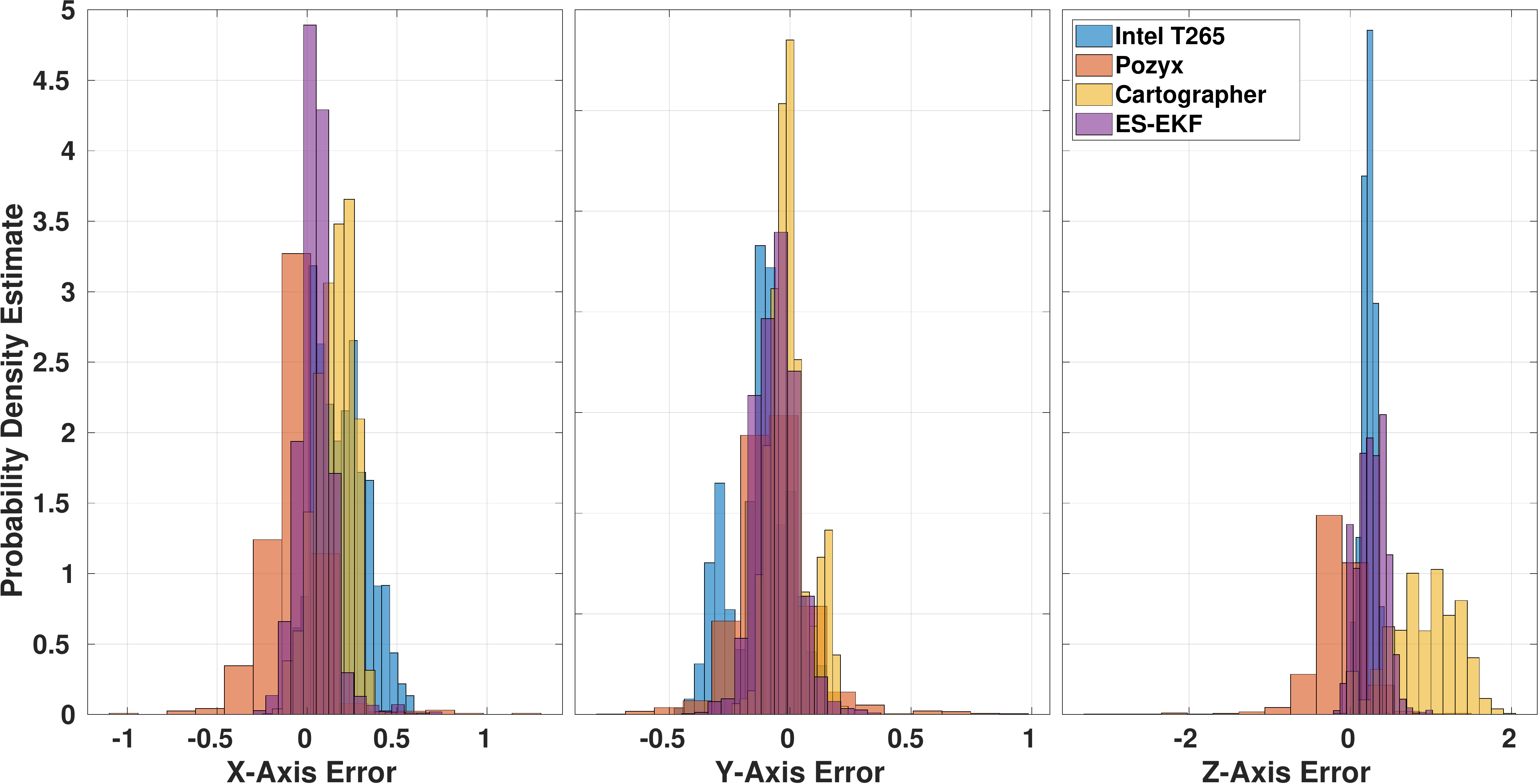}
	\caption{This figure shows position error histograms along each of the axes. The histogram values are an estimate of the underlying probability distribution of the error data. The probability distribution estimate is obtained by comparing data against a known probability density function.}
	\label{fig:histogram_comparison}
	\vspace{-0.7cm}
\end{figure}

\begin{table}[t]
    \centering
    \begin{tabular}{l*{6}{c}r}
        & & \multicolumn{2}{c}{RMSE} & & \\
        \cline{2-5}
        Estimate     & $x[\text{m}]$ & $y[\text{m}]$ & $z[\text{m}]$ & $yaw[^{\circ}]$ \\
        \toprule
        Cartographer     & 0.18 & \textbf{0.08} & 1 & 1.15 \\
        Pozyx            & 0.18 & 0.17 & 0.39 & N/A \\
        Intel T265           & 0.24 & 0.18 & \textbf{0.24} & 0.8 \\
        ES-EKF     & \textbf{0.1} & 0.1 & 0.31 & \textbf{0.49} \\
        \bottomrule
    \end{tabular}
    \caption{This table shows the root mean square error (RMSE) values for each of the position axes and yaw angle. Roll and pitch errors are sufficiently small for each sensor and are therefore omitted. It is important to note that Pozyx is not an orientation sensor.}
    \label{tab:rmse_table}
    \vspace{-0.5cm}
\end{table}

\begin{figure}[t]
	\centering
	\includegraphics[width=0.8\columnwidth]{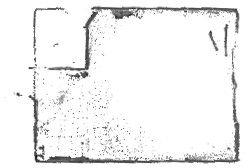}
	\caption{Map of the indoor laboratory created by Cartographer during the experimental validation. The room is 10 m by 7.5 m.}
	\label{fig:histogram_comparison}
	\vspace{-0.3cm}
\end{figure}

\begin{figure}[t]
	\centering
	\includegraphics[width=\columnwidth]{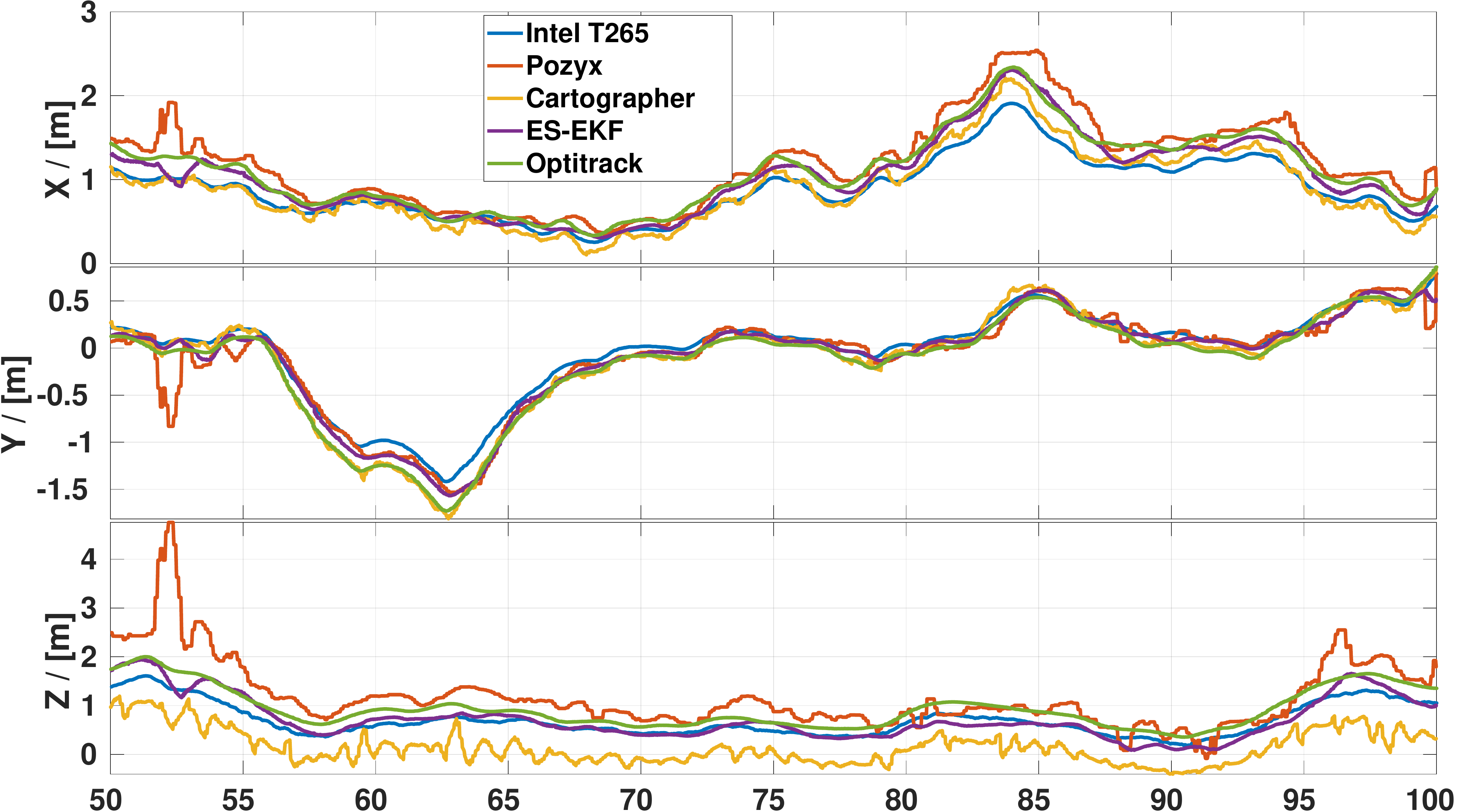}
	\caption{This figure shows the same position comparison data as Fig. \ref{fig:position_comparison} but on a smaller time-scale to better showcase the position estimate dynamics.}
	\label{fig:position_comparison_zoom}
	\vspace{-0.5cm}
\end{figure}

\begin{figure}[t]
	\centering
	\includegraphics[width=\columnwidth]{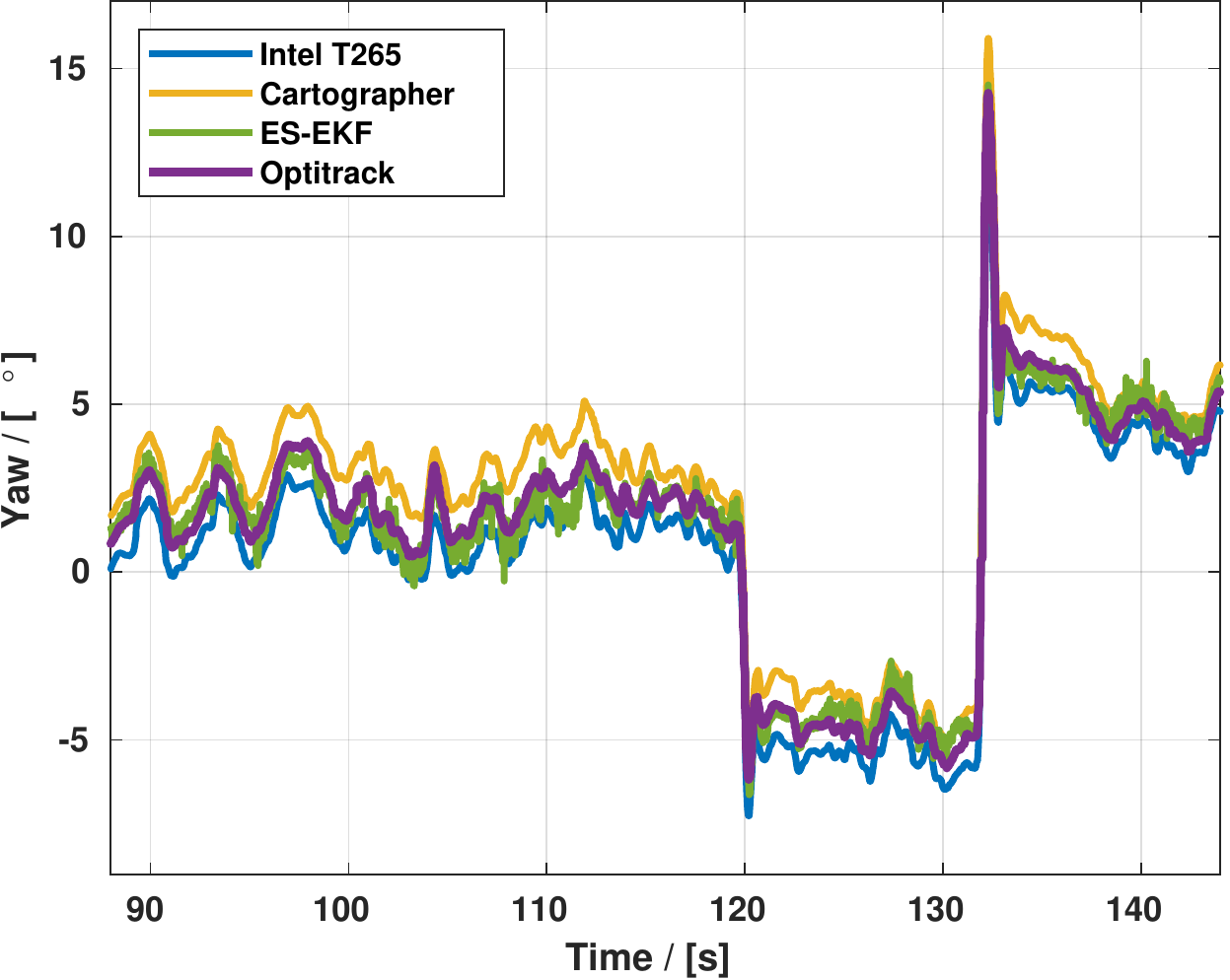}
	\caption{This figure shows a comparison of yaw estimates  between  Intel  Realsense  T265  visual inertial odometry, Cartographer SLAM, ES-EKF multi-sensor fusion and Optitrack  motion  capture  system. The data is shown on a smaller time-scale than  \ref{fig:position_comparison} in order to better showcase the estimated yaw dynamics.}
	\label{fig:yaw_comparison_zoom}
	\vspace{-0.7cm}
\end{figure}

\subsection{Discussion}

Taking into account all the data presented, namely Fig. \ref{fig:histogram_comparison} and Table \ref{tab:rmse_table}, it is safe to claim that the ES-EKF based multi-sensor fusion is able to achieve higher overall accuracy than a single sensor. Although the Intel T265 camera may be slightly more accurate along the z-axis, the Pozyx and Cartographer measurements along the same axis are far worse than the ES-EKF estimate. Altitude spikes and heavy drift can be observed in Fig. \ref{fig:position_comparison} c) or Fig. \ref{fig:histogram_comparison} - Z-axis error plot. 
Position controlled flight dynamics can be closely examined in Fig. \ref{fig:position_comparison_zoom} and Fig. \ref{fig:yaw_comparison_zoom}. The estimated UAV state indicates  no noise or delay wrt. the ground truth suggesting stable flight conditions (see \cite{videoPlaylist}).

\section{Conclusion}
\label{sec:conclusion}

In this paper an ES-EKF multi-sensor fusion method is proposed for indoor UAV localization in GPS and magnetometer denied environments. In addition to IMU data, which is indispensable, an Ouster OS1-16 LiDAR is used in combination with Cartographer SLAM to provide UAV pose measurements, an Intel RealSense T265 camera provides visual odometry information and a Pozyx UWB positioning system provides position measurements only. Since SLAM based navigation methods are known to accumulate drift over time, the ES-EKF model is extended to account for drift in sensor position and orientation. This method also helps with sensor calibration inaccuracies. The proposed state estimation system is validated in an indoor laboratory setup using the Optitrack motion capture system as the ground truth. To conclude, an ES-EKF used in the context of a multi-sensor fusion system is overall more accurate than a single-sensor state estimate. Stable flight dynamics are also demonstrated using estimated UAV state in a position control loop.

\section*{ACKNOWLEDGMENT}
This research was supported by European Commission Horizon 2020 Programme through project under G. A. number 820434, named ENergy aware BIM Cloud Platform in a COst-effective Building REnovation Context - ENCORE \cite{ENCOREweb}. Furthermore, this research was a part of the scientific project Autonomous System for Assessment and Prediction of infrastructure integrity (ASAP) \cite{ASAPweb} financed by the European Union through the European Regional Development Fund-The Competitiveness and Cohesion Operational Programme (KK.01.1.1.04.0041).






\bibliographystyle{ieeetr}
\bibliography{bibliography/references}

\end{document}